\documentclass[conference]{IEEEtran}
\IEEEoverridecommandlockouts
\usepackage[T1]{fontenc}
\usepackage{multirow}
\usepackage{tabularx,array}
\usepackage{makecell}
\usepackage{url}
\usepackage{makecell}
\renewcommand{\url}[1]{}

\usepackage{comment}
\usepackage{cite}
\usepackage{amsmath,amssymb,amsfonts}
\usepackage{algorithmic}
\usepackage{graphicx}
\usepackage{textcomp}
\usepackage{xcolor}
\def\BibTeX{{\rm B\kern-.05em{\sc i\kern-.025em b}\kern-.08em
    T\kern-.1667em\lower.7ex\hbox{E}\kern-.125emX}}
\begin{document}

\title{Embedded ConvNet Ensembles: A Lightweight Approach to Recognize Arabic Handwritten Characters\\
}

\author{\IEEEauthorblockN{1\textsuperscript{st} Mohsine El Khayati}
\IEEEauthorblockA{\textit{Systems theory and informatics laboratory} \\
\textit{Moulay Ismail University of Meknes}\\
Meknes, Morocco \\
m.elkhayati@umi.ac.ma}
\and
\IEEEauthorblockN{2\textsuperscript{nd} Rachid Elouahbi}
\IEEEauthorblockA{\textit{Laboratory of Computer Science and Applications} \\
\textit{Moulay Ismail University of Meknes}\\
Meknes, Morocco \\
r.elouahbi@umi.ac.ma}
\and
\IEEEauthorblockN{3\textsuperscript{rd} Abdelillah Semma}
\IEEEauthorblockA{\textit{Computer Science Dept.} \\
\textit{EST of Sidi Bennour}\\
\textit{Chouaib Doukkali University}\\
Eljadida, Morocco}
}

\maketitle

\begin{abstract}
Arabic Handwritten Character Recognition (AHCR) has recently advanced significantly with deep Convolutional Neural Networks (ConvNets). However, many models in the literature are deep and computationally expensive in terms of parameters and FLOPs, limiting their deployment on resource-constrained devices, which are increasingly common. This study addresses this gap by proposing a combination of lightweight embedded ConvNet models and ensemble learning techniques. Extensive experiments were conducted to identify best practices in AHCR, considering training hyperparameters, learning strategies, model choices, and ensemble methods. Results show that embedded models can achieve accuracy comparable to, or even surpassing, heavier architectures. Ensemble learning further enhances performance with only modest computational overhead, particularly under challenging training scenarios. Among the ensembling strategies, soft voting yielded the best overall results.
\end{abstract}


\section{Introduction}
Arabic Handwriting Recognition (AHR) refers to the process of transforming images of handwritten Arabic text into a machine-readable digital format. AHR has numerous real-world applications, including banking, the archiving of historical documents, information retrieval and indexing, and human–computer interaction \cite{Lamaakal2024}. Research in this area is advancing, and recent results have been increasingly satisfactory in various tasks such as character, digit, and word recognition. However, to our knowledge, universal solutions for recognizing entire texts are still missing. This is primarily due to challenges inherent to the Arabic script, such as its cursive nature and calligraphic styles \cite{ElHajj2005}. 

Two main approaches to address AHR are widely adopted in the literature: the holistic approach and the analytic approach \cite{ElKhayati2024}. In the holistic approach \cite{Alnuzaili2012}, the words are recognized as single units, whereas in the analytic approach \cite{Eraqi2016}, the word is segmented into characters, which are then recognized separately. Within the analytic approach, the recognition of characters constitutes a crucial step, commonly referred to in the literature as Arabic Handwritten Character Recognition (AHCR). 

Recently, deep learning techniques, particularly the Convolutional Neural Networks (ConvNet) family, have driven significant advances in AHCR \cite{Alwagdani2023,Alyahya20,altwaijry2021,ElKhayati2024_overview,ElKhayati2025_within_domain}. Recognition accuracies have improved significantly compared to traditional classification techniques \cite{Wagaa2022, Alwagdani2023, Alheraki2023}. However, most existing AHCR studies focus on maximizing accuracy using heavy ConvNet models such as AlexNet \cite{Almodfer2017}, ResNet \cite{Taani2021}, and VGGNet \cite{Lamtougui2024}. Although effective, these models have high parameter counts and computational costs, limiting their applicability on mobile and embedded devices \cite{Lamaakal2024,ElKhayati2025_laveraging}. Meanwhile, ensemble learning has shown promise in improving performance in various AHR tasks \cite{Nanehkaran21, Sousa18, Ahranjany10}, but its use with lightweight models remains underexplored. There is a notable gap in the literature regarding the evaluation of mobile-friendly models and their combination through ensemble learning for AHCR. 

This study addresses the gap by introducing four embedded ConvNet models (MobileNet, SqueezeNet, ShuffleNet, and MnasNet) to AHCR and combining their outcomes using three different ensemble learning strategies. Extensive experiments were conducted on three benchmark datasets—AHCD, Hijja, and IFHCDB— to identify best practices regarding training hyperparameters, model selection, ensembling strategies, and transfer learning approaches.

The remainder of this paper is organized as follows: Section 2 reviews related works. Section 3 presents the proposed method. Section 4 reports and analyzes the results. Section 5 discusses the findings. Finally, Section 6 concludes the paper.

\section{Related works}
Ensemble learning involves aggregating the outputs of multiple classifiers. In this work, we focus on studies that apply ensemble learning with ConvNets in the context of AHR, which aligns with our approach.

De Sousa \cite{Sousa18} applied ensemble learning to four models for Arabic digit and character recognition. The outputs of the four models were averaged, significantly improving results compared to base models, achieving 99.74\% accuracy on MADBase (digits) and 98.42\% on AHCD (isolated characters).

Alyahya et al. \cite{Alyahya20} investigated ensemble learning on the standard version of ResNet-18 and a modified version incorporating dropout layers after convolutional layers. Both models were trained with and without data augmentation. Contrary to the findings of \cite{Sousa18}, base models surpassed the ensemble’s performance.

For word recognition, Awni et al. \cite{Awni19} trained ResNet-18 with three distinct optimizers and combined their predictions via averaging. The ensemble surpassed the base models, attaining a 6.63\% error rate on IFN/ENIT.

A similar strategy was used by Almodfer et al. \cite{Almodfer17}, varying image resolutions instead of optimizers. They trained the same ConvNet with three different resolutions: 100 px, 200 px, and 300 px. The ensemble model outperformed base models, achieving a 2.4\% lower word error rate than the best single model.

Ahranjany et al. \cite{Ahranjany10} employed ensemble learning for digit recognition by aggregating outputs from five LeNet-5 models using four methods: average, maximum, minimum, and product. Their experiments showed that averaging provided the best performance across all tests, resulting in the lowest digit recognition error. The approach attained 99.17\% accuracy on IFHCDB and 99.98\% after rejecting hard-to-recognize samples. 
 
\section{Method}

\begin{figure}
    \centering
    \includegraphics[height=0.4\textheight]{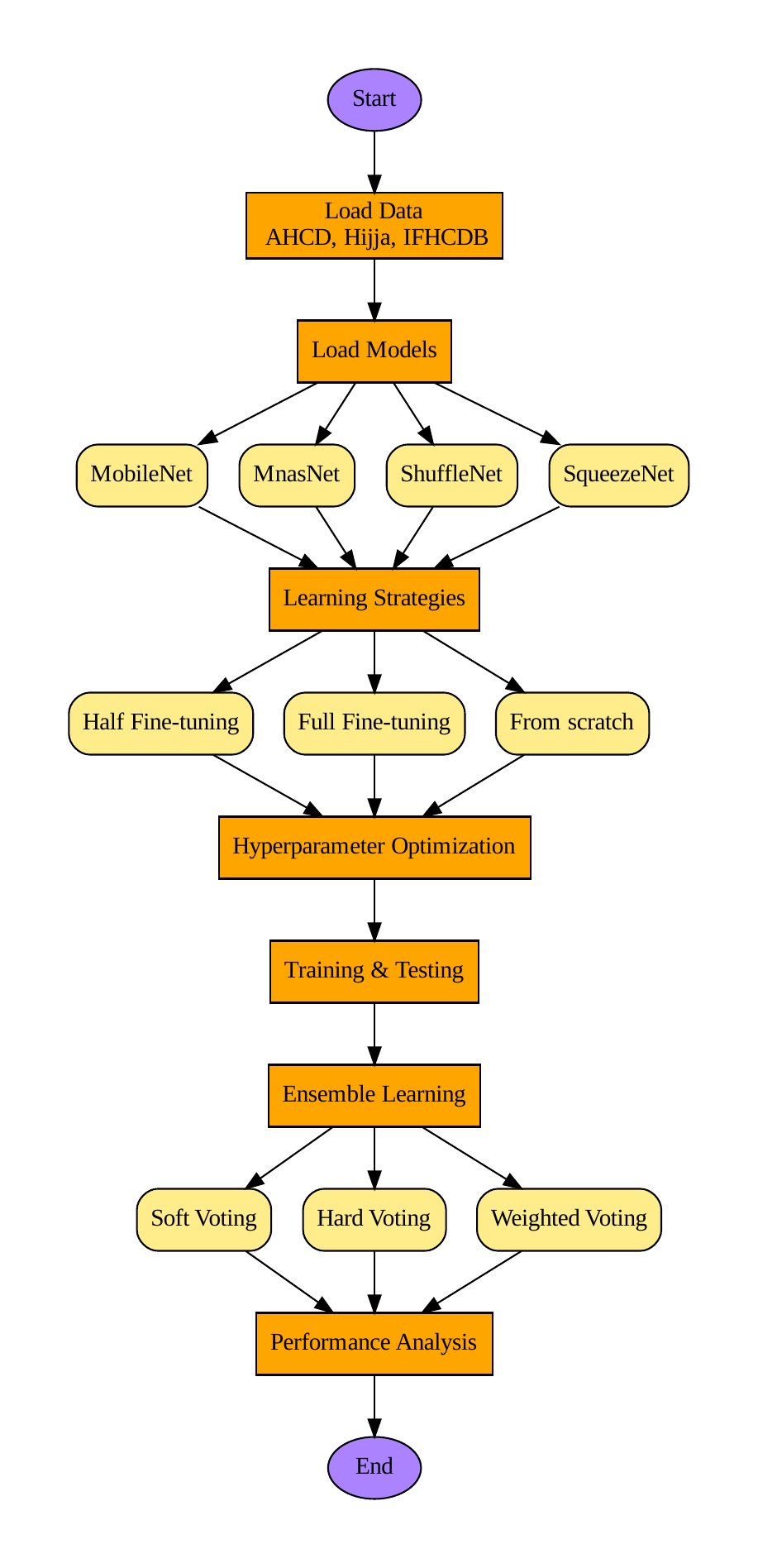}
    \caption{Flowchart of the proposed method}
    \label{fig:method}
\end{figure}

This study examines the effect of mobile-friendly ConvNet architectures and ensemble learning within AHCR. The adopted methodology is depicted in Fig. \ref{fig:method}.

First, three benchmark datasets were collected for experiments: the Isolated Farsi Handwritten Character Dataset (IFHCDB) \cite{mozaffari2006}, the Arabic Handwritten Character Dataset (AHCD) \cite{elsawy2017}, and Hijja \cite{altwaijry2021}. These datasets were chosen due to their extensive adoption in the literature, enabling direct comparison of results. The characteristics of the datasets are summarized in Table \ref{tab:database}.

Next, four models were selected for their lightweight architectures: MobileNet \cite{howard17,sandler2018}, MnasNet \cite{tan19}, ShuffleNet \cite{zhang18}, and SqueezeNet \cite{iandola16}. These models have minimal parameter counts and the number of Floating Point Operations (FLOPs), making them suitable for mobile and IoT devices. They were originally designed to operate efficiently in resource-constrained environments while maintaining performance comparable to heavier models. Table \ref{tab:model_comparison} provides empirical justification for the selection of the examined models, highlighting their markedly lower parameter counts and FLOPs in comparison with state-of-the-art counterparts.

Three learning strategies were employed: Training From Scratch (TFS), Half Fine-tuning (HFT), and Full Fine-tuning (FFT). In TFS, models are trained from random weight initialization. In HFT, weights of the feature extraction layers—pretrained on ImageNet—are frozen, and only the classification layers are fine-tuned. In FFT, pretrained weights are used as initializers, and all layers are fine-tuned.

A hyperparameter optimization step was performed before training to identify the best hyperparameters. The Hyperband HPO strategy \cite{li2018hyperband} was applied, known for its efficiency in eliminating non-promising trials. This method allows running multiple trials in parallel, dynamically replacing failing ones, enabling an efficient expansion of the search space, and maximizing hyperparameter exploration. The technique was applied across all models, training strategies, and datasets (36 combinations). The best-performing hyperparameter sets were saved for use during training.

Finally, three ensembling strategies were evaluated: soft voting, hard voting, and weighted voting. In soft voting, model output probabilities are aggregated, and the class with the highest combined probability is selected. Weighted voting follows the same process but assigns weights to base model outputs based on their performance, favoring stronger models. Each model’s contribution is proportional to its normalized performance score. Specifically, weights are obtained by normalizing validation scores across models, ensuring that higher-performing models have a greater influence on the final aggregated prediction. In hard voting, the class with the majority of votes from base models' predictions is selected. Various model combinations were tested, as detailed in Section \ref{ens_str_comp}.

Training and testing were performed using the original dataset splits described in Table \ref{tab:database}. During training, 5-fold cross-validation was applied, where the training set is divided into five equal folds: four for training and one for validation. This process is repeated five times so that each fold serves as validation once. The best-performing fold is then selected for final testing. All models were evaluated individually on the testing set. To enhance reproducibility and flexibility, model weights were saved after each epoch in separate .pt files, enabling training or testing to resume from any checkpoint when needed. During inference, the checkpoint corresponding to the highest validation accuracy is loaded for evaluation on the test set.

\begin{table}
\caption{Overview of dataset characteristics used in the study. Only Arabic characters from IFHCDB were retained.}
\label{tab:database}
\centering
\resizebox{\columnwidth}{!}{
\begin{tabular}{|c|c|c|c|c|c|}
\hline
Dataset & Dataset size & Train set & Test set & Images size \\
\hline
IFHCDB \cite{mozaffari2006} & 70,120 &  36,017 (75\%)& 12,440 (25\%) & $77\times95$  \\
AHCD \cite{elsawy2017} & 16,800 & 13,440 (80\%) & 3,360 (20\%)& $32\times32$  \\
Hijja \cite{altwaijry2021} & 47,434 & 37,933 (80\%) & 9,500 (20\%) & $32\times32$\\
\hline
\end{tabular}
}
\end{table}

\begin{table}[t]
\caption{Comparison of the studied models against literature models in terms of parameters and FLOPs. Parameters in Millions}
\label{tab:model_comparison}
\centering
\begin{tabular}{|p{4cm}|p{1.2cm}|p{1cm}|}
\hline
Models & Parameters & GFLOPs \\
\hline
\multicolumn{3}{|c|}{Top used models for AHR} \\
\hline
AlexNet \cite{9,arif2020} & 61 & 0.72 \\
ResNet50 \cite{6,7,8} & 25 & 4 \\
VGG16 \cite{6,7} & 138 & 16 \\
VGG19 \cite{7} & 143 & 20 \\
Inception V3 \cite{6,7} & 27 & 6 \\
\hline
\multicolumn{3}{|c|}{The models utilized in this work} \\
\hline
MobileNetV3\_small weights \cite{howard17} & 2.5 & 0.02 \\
ShuffleNet\_V2\_X0\_5 \cite{zhang18} & 1.4 & 0.013 \\
MnasNet0\_5 \cite{tan19} & 2.2 & 0.31 \\
SqueezeNet1\_1 \cite{iandola16} & 1.2 & 0.36 \\
\hline
\end{tabular}
\end{table}

\section{Results}

\subsection{Testing performance of base models}
\label{sec:ind}

Table \ref{tab:ind_perf} summarizes the testing performance of the studied base models under different learning strategies. Several conclusions can be drawn from the results. MobileNet and ShuffleNet showed greater stability across datasets and training strategies. SqueezeNet and MnasNet experienced overfitting issues, particularly under the HFT strategy. The results further indicate that learning strategies significantly impact results, with FFT and TFS achieving higher performance, while HFT underperforms in most experiments. This is mainly due to the pretrained weights originating from a different domain (the ImageNet dataset for natural images). Regarding the datasets, IFHCDB yielded the highest accuracy across most experiments, whereas Hijja produced the lowest performance, with Many models encountered difficulties in convergence and generalization on Hijja.

\begin{table*}[htbp]
\centering
\caption{Testing performance of the base models on three benchmark datasets under various learning strategies.}
\label{tab:ind_perf}
\begin{tabular}{|c|c|c|c|c|c|c|c|c|c|c|c|c|c|c|c|c|}
\hline

\multirow{2}{*}{Dataset} & \multirow{2}{*}{Model} & \multicolumn{4}{|c|}{TFS} & \multicolumn{4}{|c|} {HFT} & \multicolumn{4}{|c|}{FFT}\\
\cline{3-14}
& &Acc & F1 & Prec & Rec &Acc & F1 & Prec & Rec &Acc & F1 & Prec & Rec\\
\hline

AHCD & MobileNet &0.960&0.960&0.960&0.957 &0.737&0.750&0.737&0.724 &0.96&0.961&0.96&0.956\\

& MnasNet &0.903&0.917&0.903&0.904 &0.262&0.366&0.262&0.220 &0.727&0.85&0.727&0.711 \\

& ShuffleNet &0.967&0.97&0.967&0.967 &0.804&0.782&0.804&0.791 &0.97&0.971&0.97&0.97\\

& SqueezeNet &0.96&0.961&0.96&0.96 &0.820&0.832&0.820&0.810 &0.954&0.96&0.954&0.952\\

\hline
Hijja & MobileNet &0.897&0.897&0.897&0.897 &0.482&0.48&0.482&0.48 &0.903&0.903&0.903&0.903\\

& MnasNet &0.901&0.902&0.901&0.901&0.431&0.43&0.431&0.425 &0.85&0.85&0.85&0.85\\

& ShuffleNet &0.912&0.913&0.912&0.912 &0.536&0.54&0.536&0.535 &0.92&0.92&0.92&0.92\\

& SqueezeNet &0.882&0.885&0.882&0.882 &0.474&0.494&0.474&0.476 &0.911&0.912&0.911&0.911\\
\hline

IFHCDB & MobileNet &0.98&0.98&0.98&0.98 &0.912&0.91&0.912&0.91 &0.983&0.983&0.983&0.983\\
& MnasNet &0.982&0.983&0.982&0.982 &0.376&0.45&0.376&0.274 &0.986&0.987&0.986&0.986\\

& ShuffleNet &0.975&0.975&0.975&0.975 &0.892&0.89&0.892&0.885 &0.98&0.98&0.98&0.98\\

& SqueezeNet &0.981&0.981&0.981&0.981 &0.952&0.953&0.952&0.951 &0.983&0.983&0.983&0.983\\
\hline
\end{tabular}
\end{table*}

\subsection{Ensemble learning strategies comparison} \label{ens_str_comp}

Tables \ref{tab:el_ahcd}, \ref{tab:el_hijja}, and \ref{tab:el_ifhcdb} present the performance of various ensemble learning combinations on the AHCD, Hijja, and IFHCDB datasets, respectively. Multiple model combinations were evaluated to determine optimal strategies. The following explains the combinations detailed in these tables.

\begin{itemize}
    \item \textbf{All-Ens:} the complete set of 12 models trained on the same dataset.
    \item \textbf{TFS-Ens:} models trained exclusively with the TFS strategy (4 models); similarly for HFT-Ens and FFT-Ens.
    \item \textbf{MobileNet-Ens:} ensemble of MobileNet models trained under different strategies (3 models); similarly for other models.
    \item \textbf{Best-Ens:} Base models exhibiting higher performance were selected, while lower-performing models were excluded.
\end{itemize}

 From Tables \ref{tab:el_ahcd}, \ref{tab:el_hijja}, and \ref{tab:el_ifhcdb}, it is observed that the Best-Ens outperformed other combinations in the majority of experiments. However, some combinations yielded comparable results under specific conditions; for instance, the All-Ens attained an accuracy of 0.989 on IFHCDB with both soft and weighted voting, while TFS-Ens achieved 0.972 accuracy on AHCD using soft voting. These findings align with expectations, as ensembles of top-performing base models are naturally more effective.

For the ensembling strategies, we observe that soft voting and weighted voting outperform hard voting. The performance gap is more pronounced with underperforming models such as MnasNet-Ens and HFT-Ens, with accuracy differences ranging from 2\% to 11\%. For high-performing models, hard voting achieves comparable results.

\begin{table*}[htbp]
\centering
\caption{Performance of different ensemble combinations on AHCD. \emph{ Acc: Accuracy; F1: F1 Score; Prec: Precision; Rec: Recall}}\label{tab:el_ahcd}

\begin{tabular}{|c|c|c|c|c|c|c|c|c|c|c|c|c|c|c|c|c|}
\hline

\multirow{2}{*}{\makecell{Combination}} & \multicolumn{4}{|c|}{Soft voting} & \multicolumn{4}{|c|} {Hard voting} & \multicolumn{4}{|c|}{Weighted voting}\\
\cline{2-13}
&Acc & F1 & Prec & Rec &Acc & F1 & Prec & Rec &Acc & F1 & Prec & Rec\\
\hline
All-Ens &0.969&0.968&0.972&0.969&0.968&0.967&0.970&0.968&0.971&0.970&0.973&0.971\\
TFS-Ens &0.972&0.972 &0.973&0.972&0.968&0.968&0.970&0.968&0.970  &0.969&0.971&0.970\\
HFT-Ens&0.854&0.840&0.829&0.854&0.820 &0.807&0.801&0.820&0.847&0.833&0.821&0.847\\
FFT-Ens&0.968 &0.967&0.971&0.968&0.967&0.965&0.970&0.967&0.967 &0.966&0.969&0.967\\
MobileNet-Ens &0.964 &0.962&0.967&0.964&0.956&0.954&0.961&0.956&0.964  &0.962&0.967&0.964\\
MnasNet-Ens &0.877 &0.873&0.912&0.877&0.765&0.756&0.849&0.765&0.879  &0.875&0.913&0.879\\
ShuffleNet-Ens &0.970&0.969&0.972&0.970&0.967&0.966&0.969&0.967&0.970&0.969&0.972&0.970\\
SqueezeNet-Ens &0.960&0.959&0.962&0.960&0.954&0.951&0.958&0.954&0.960  &0.959&0.963&0.960\\
Best-Ens &0.972&0.972&0.975&0.972&0.974&0.973&0.975&0.974&0.971 &0.970&0.973&0.971\\
\hline
\end{tabular}
\end{table*}

\begin{table*}[htbp]
\caption{Performance of different ensemble combinations on Hijja dataset}\label{tab:el_hijja}
\centering
\begin{tabular}{|c|c|c|c|c|c|c|c|c|c|c|c|c|c|c|c|c|}
\hline

\multirow{2}{*}{\makecell{Combination}} & \multicolumn{4}{|c|}{Soft voting} & \multicolumn{4}{|c|} {Hard voting} & \multicolumn{4}{|c|}{Weighted voting}\\
\cline{2-13}
&Acc & F1 & Prec & Rec &Acc & F1 & Prec & Rec &Acc & F1 & Prec & Rec\\
\hline
All-Ens&0.930&0.930&0.930&0.930 &0.926&0.926&0.927&0.926&0.931&0.931&0.932&0.931\\
TFS-Ens &0.924&0.924&0.925&0.924&0.919&0.919&0.920&0.919&0.925&0.925&0.925&0.925\\
HFT-Ens &0.603&0.601&0.603&0.603&0.560&0.557&0.565&0.560&0.603&0.602&0.603&0.603\\
FFT-Ens &0.927&0.927&0.927&0.927&0.922&0.922&0.922&0.922&0.928&0.928&0.928&0.928\\
MobileNet-Ens&0.912&0.912&0.913&0.912&0.894&0.894&0.896&0.894&0.914&0.914&0.914&0.914\\
MnasNet-Ens&0.896&0.896&0.897&0.896&0.860&0.859&0.863&0.860&0.902&0.902&0.902&0.902\\
ShuffleNet-Ens&0.925&0.925&0.925&0.925&0.913&0.912&0.913&0.913&0.925&0.925&0.925&0.925\\
SqueezeNet-Ens&0.912&0.912&0.913&0.912&0.891&0.891&0.893&0.891&0.914&0.914&0.915&0.914\\
Best-Ens &0.933&0.933&0.933&0.933&0.931&0.931&0.932&0.931&0.933&0.933&0.934&0.933\\
\hline
\end{tabular}
\end{table*}

\begin{table*}[htbp]
\caption{Performance of different ensemble Combinations on IFHCDB dataset}\label{tab:el_ifhcdb}
\centering
\begin{tabular}{|c|c|c|c|c|c|c|c|c|c|c|c|c|c|c|c|c|}
\hline

\multirow{2}{*}{\makecell{Combination}} & \multicolumn{4}{|c|}{Soft voting} & \multicolumn{4}{|c|} {Hard voting} & \multicolumn{4}{|c|}{Weighted voting}\\
\cline{2-13}
&Acc & F1 & Prec & Rec &Acc & F1 & Prec & Rec &Acc & F1 & Prec & Rec\\
\hline
All-Ens&0.989&0.989&0.989&0.989&0.988&0.988&0.988&0.988&0.989&0.989&0.989&0.989\\
TFS-Ens &0.985&0.985&0.985&0.985&0.984&0.984&0.984&0.984&0.985&0.985&0.985&0.985\\
HFT-Ens &0.953&0.950&0.953&0.953&0.935&0.931&0.933&0.935&0.955&0.952&0.955&0.955\\
FFT-Ens &0.990&0.990&0.990&0.990&0.989&0.989&0.989&0.989&0.990&0.990&0.990&0.990\\
MobileNet-Ens&0.984&0.984&0.984&0.984&0.981&0.981&0.981&0.981&0.984&0.984&0.984&0.984\\
MnasNet-Ens&0.988&0.988&0.988&0.988&0.983&0.983&0.984&0.983&0.988&0.988&0.988&0.988\\
ShuffleNet-Ens&0.979&0.979&0.980&0.979&0.976&0.975&0.976&0.976&0.980&0.980&0.980&0.980\\
SqueezeNet-Ens&0.986&0.986&0.986&0.986&0.984&0.984&0.984&0.984&0.986&0.986&0.986&0.986\\
Best-Ens &0.989&0.989&0.989&0.989&0.989&0.989&0.989&0.989&0.989&0.989&0.989&0.989\\
\hline
\end{tabular}
\end{table*}

\subsection{Base versus Ensemble Performance}

Table \ref{tab:ind_ens} presents a comparison between the best base model accuracies and those achieved by ensemble learning techniques. The results show that ensemble learning outperformed the best base models by 0.4\% on AHCD, 1.3\% on Hijja, and 0.4\% on IFHCDB. Notably, ensemble learning surpassed all base models, with the performance gap being more pronounced under challenging conditions, such as on the Hijja dataset.

\begin{table}
\centering
\caption{Comparison between top-performing base models and ensemble learning accuracies. \emph{Bold indicates top results. }}\label{tab:ind_ens}
\begin{tabular}{|c|c|c|c|}
\hline
Model & AHCD & Hijja & IFHCDB \\
\hline
MobileNet & 0.958	& 0.903 & 0.982\\
MnasNet&0.903&0.901&0.986\\ShuffleNet&0.970&0.920&0.980\\SqueezeNet&0.960&0.911&0.982\\Ensemble& \textbf{0.974} & \textbf{0.933} & \textbf{0.990}\\
\hline
\end{tabular}
\end{table}

\subsection{Computational Complexity}

All experiments were conducted on Google Colab using a virtual machine equipped with an Intel(R) Xeon(R) CPU @ 2.20GHz, featuring two cores and support for 64-bit operations. The system provided 12 GB of RAM and approximately 108 GB of disk space. All computations were executed in CPU-only mode to approximate resource-constrained conditions.

Table \ref{tab:complexity} presents the computational performance of the base models and two ensemble configurations. The first ensemble  (\emph{Ensemble-4}) includes all four base models, while the second (\emph{Ensemble-3}) excludes SqueezeNet, which showed notably higher latency and inference time. The individual lightweight architectures—MobileNet, MnasNet, and ShuffleNet—demonstrate low latency and fast inference. SqueezeNet, however, stands out with substantially higher latency (2.282 s) and inference time (0.061 s).

The full four-model Ensemble-4 (including SqueezeNet) significantly increases latency (3.435 s) and inference time (0.102 s), showing that SqueezeNet is the primary contributor to the computational overhead. When SqueezeNet is excluded, the three-model Ensemble-3 reduces latency to 1.041 s and inference time to 0.044 s—cutting the total delay by more than two-thirds—while maintaining nearly identical load and inference memory.

\begin{table}
\centering
\caption{Comparison between base and ensemble models in terms of latency, inference time, and memory load. \emph{time in seconds and memory in MB}}
\label{tab:complexity}
\begin{tabular}{|c|c|c|c|c|}
\hline
Model & 
\makecell{Latency\\(batch)} & 
\makecell{Inference\\time (img)} & 
\makecell{Load\\memory} & 
\makecell{Inference\\memory (img)} \\
\hline

  MobileNet & 0.240 & 0.014 & 21.50 & 0.12\\
  MnasNet & 0.546 & 0.017 & 20.12 & 0.00\\
  ShuffleNet &  0.333 & 0.016 & 7.87 & 1.00\\
  SqueezeNet & 2.282 & 0.061 & 13.25 & 5.12\\
Ensemble-4 & 3.435 & 0.102 &35.50 & 0.12\\
Ensemble-3 & 1.041 & 0.044 &34.87 & 0.12\\
\hline
\end{tabular}
\end{table}

\subsection{Comparison with the literature}
Table \ref{tab:comp_liter} presents a comparison between the highest testing accuracies achieved in this study and those reported in related literature. To ensure fairness, only studies closely matching our datasets, data splits, and class counts were considered. Studies involving data augmentation were omitted.

The proposed approach achieved the highest accuracy on all datasets, surpassing existing works in the literature. The improvement on AHCD is modest, with \cite{Ullah2022} and \cite{Alheraki2023} reporting comparable results. However, the gap is more pronounced on Hijja and IFHCDB, where our method outperformed others by up to 1.4\%. 

\begin{table}[t]
\centering
\caption{Comparison of the best obtained accuracies with those reported in recent literature}
\label{tab:comp_liter}
\resizebox{\columnwidth}{!}{%
\begin{tabular}{|c|c|c|c|c|}
\hline
Author & Model & AHCD & Hijja & IFHCDB\\
\hline
Ullah et al. \cite{Ullah2022} & Custom ConvNet & 0.967 & - & - \\
Alheraki et al. \cite{Alheraki2023} & Custom ConvNet & 0.970 & 0.910 & - \\
Alwagdani et al. \cite{Alwagdani2023} & Hybrid CNN-SVM & - & 0.919 & - \\
Dhief et al. \cite{Dhief2022} & Fusion of classifiers' decisions & - & - & 0.979 \\
Arif and Poruran \cite{arif2020} & OCR-AlexNet & - & - & 0.960\\
Arif and Poruran \cite{arif2020} & OCR-GoogleNet & - & - & 0.940\\
Proposed method & Ensemble embedded ConvNets & \textbf{0.974} & \textbf{0.933} & \textbf{0.990}\\
\hline
\end{tabular}}
\end{table}

\section{Discussion}
The results indicate that ensemble learning improves performance, with accuracy gains ranging from 0.4\% to 1.4\% compared to base models. The benefit of ensemble learning is especially pronounced on the Hijja dataset, which is a complex dataset. The substantial improvements achieved by ensemble learning suggest it is a promising solution for handling such complex datasets.

Soft voting and weighted voting have proven more effective than hard voting in most experiments. Their advantage is especially pronounced when the ensemble includes underperforming models, whereas Hard voting achieves similar effectiveness when most base models demonstrate high accuracy. These results suggest that soft and weighted voting are preferable when ensembles contain lower-confidence models, though hard voting remains a viable option if models' performance is well calibrated.

The findings of this study confirm the efficacy of embedded models, surpassing heavier architectures like AlexNet and GoogleNet \cite{arif2020} reported in the literature. This underscores the promise of lightweight models for AHCR and encourages further exploration. Ensemble learning similarly demonstrated effectiveness in enhancing model outcomes under diverse training conditions, especially difficult scenarios, aligning with prior research \cite{Sousa18,Awni19,Almodfer17}.

The computational analysis highlights that model selection within an ensemble critically affects efficiency. SqueezeNet, while compact in design, introduces disproportionate latency and inference time, which substantially increases the ensemble’s overall complexity. Naturally, an ensemble is more resource-demanding than a single model, but it delivers measurable performance gains, particularly on complex datasets such as Hijja.

As a limitation of this study, we note the blind training. Although the results were satisfactory, architectural improvements and data pre-processing could enhance performance and/or reduce computational complexity. A detailed investigation into the architectural impact is thus necessary. Additionally, the study does not explain why the same models perform differently across datasets. While this is attributed to dataset complexity, a more thorough analysis is required to understand the behavioral differences of each model.

\section{Conclusion}
This study focused on enhancing AHCR with attention to computational efficiency. To achieve this, we investigated a combination of embedded ConvNet models and ensemble learning techniques through extensive experiments and comparisons. The results demonstrate that embedded models are effective, and ensemble learning enhances performance, particularly under challenging training scenarios. Soft and weighted voting are more suitable for ensembles with less confident models, whereas hard voting is effective when models exhibit high confidence. The computational analysis further revealed that the ensemble’s complexity remains moderate and can be significantly reduced by excluding inefficient models such as SqueezeNet. Even in its full configuration, the ensemble remains computationally lighter than a single heavy architecture, while offering superior accuracy, reinforcing its practicality for real-world applications.

Future work will focus on studying the robustness, sensitivity, and calibration of the models under challenging training and inference conditions. Additionally, the computational complexity of the proposed method will be evaluated and compared with well-known ConvNet architectures.

\bibliographystyle{IEEEtran}
\bibliography{references}

\end{document}